# Is the deconvolution layer the same as a convolutional layer?

A note on Real-Time Single Image and Video Super-Resolution Using an Efficient Sub-Pixel Convolutional Neural Network.

Wenzhe Shi, Jose Caballero, Lucas Theis, Ferenc Huszar, Andrew Aitken, Alykhan Tejani, Johannes Totz, Christian Ledig, Zehan Wang

Twitter, Inc.[1]

In our CVPR 2016 paper [1], we proposed a novel network architecture to perform single image super-resolution (SR). Most existing convolutional neural network (CNN) based super-resolution methods [10,11] first upsample the image using a bicubic interpolation, then apply a convolutional network. We will refer to these types of networks as high-resolution (HR) networks because the images are upsampled first. Instead, we feed the low-resolution (LR) input directly to a sub-pixel CNN as shown in Fig.1:

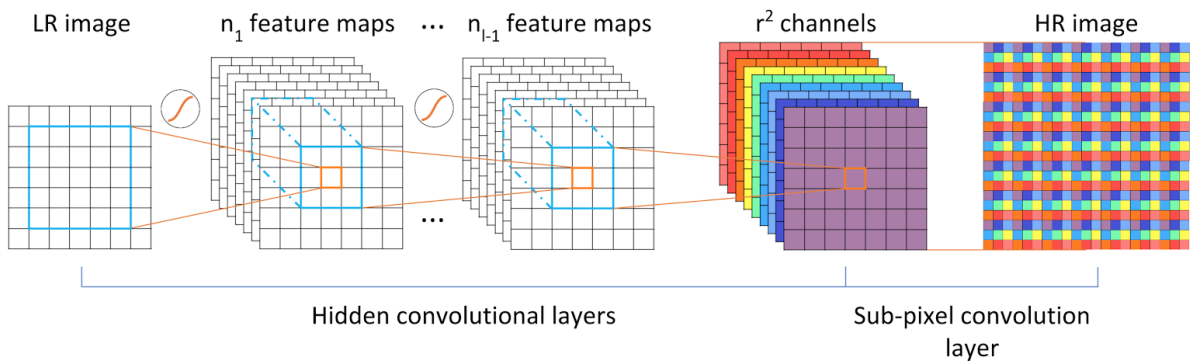

Figure 1: An illustration of the ESCPN framework where $r$ denotes the upscaling ratio.

Let $r$ denote the upscaling ratio - e.g if the input LR image is $1 \times 1$ then the output HR image will be $r \times r$. We then output $r^2$ number of channels instead of one high-resolution (HR) image and use periodic shuffling to recreate the HR image. The exact details about how our efficient sub-pixel convolutional layer works can be found in the paper. We will refer to our network as a LR network.

In this note, we want to focus on two aspects related to two questions most people asked us at CVPR when they saw this network. Firstly, how can $r^2$ channels magically become a HR image? And secondly, why are convolution in LR space a better choice? These are actually the key questions we tried to answer in the paper, but we were not able to go into as much depth and clarity as we would've liked given the page limit. To better answer these questions, we first discuss the relationships between the deconvolution layer in the form of the transposed convolution layer, the sub-pixel convolutional layer and our efficient sub-pixel convolutional layer, which we'll go through in Sec. 1 and Sec. 2. We will refer to our efficient sub-pixel convolutional layer as a convolutional layer in LR space to distinguish it from the common sub-pixel convolutional layer [5]. We will then show that for a fixed computational budget and complexity, a network with convolutions exclusively in LR space has more representation power at the same speed than a network that first upsamples the input in HR space.

**Section 1: Transposed convolution and sub-pixel convolutional layers**

First we need to examine the deconvolution layer. The deconvolution layer, to which people commonly refer, first appears in Zeiler's paper as part of the deconvolutional network [2] but does not have a specific name. The term deconvolution layer is used in his later work [3] and then implemented in caffe.[2]

After the success of the network visualization paper [4] it became widely adopted and is now commonly used in the context of semantic segmentation [5], flow estimation [6] and generative modeling [7]. It also has many names including (but not limited to) sub-pixel or fractional convolutional layer [7], transposed convolutional layer [8,9], inverse, up or backward convolutional layer [5,6].[3] To explain the relationships between these different names, let's start with a simple convolution with stride 2 in 1D as shown in Fig.2, which is inspired by [8,9]:

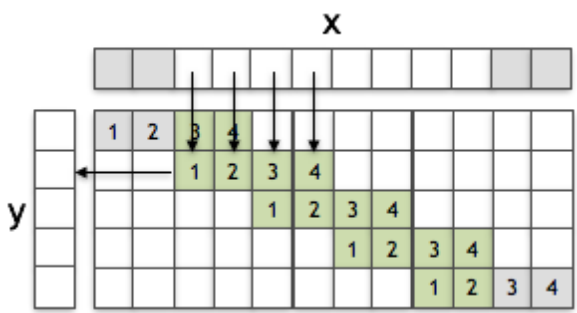

Figure 2: Convolution with stride 2 in 1D

Fig.2 illustrates 1D padded convolution of a 1D signal $x$ by a filter $f$ to obtain a 1D signal $y$. The signal $x$ is of size 8, the filter $f$ is of size 4 and the signal $y$ is of size 5. The grey areas in $x$ represent padding with zeros. The grey areas in $f$ represent multiplication with zeros. The values of $x$ that contribute to values of $y$ are shown with arrows. We note that a convolution with stride 2 is a downsampling operation.

Now, let's examine a cropped transposed convolution with stride 2 and sub-pixel convolution with stride ½ both in 1D:

---

[2] http://caffe.berkeleyvision.org/doxygen/classcaffe_1_1DeconvolutionLayer.html
[3] If the layer remembers the max pooling indices, and use the indices in the unpooling stage as in the original paper [2], then this specific form of sub-pixel convolutional layer (deconvolution layer with memorial unpooling) is distinct from the transposed convolutional layer. The rest of this note assumes that we do not memorize the pooling indices.

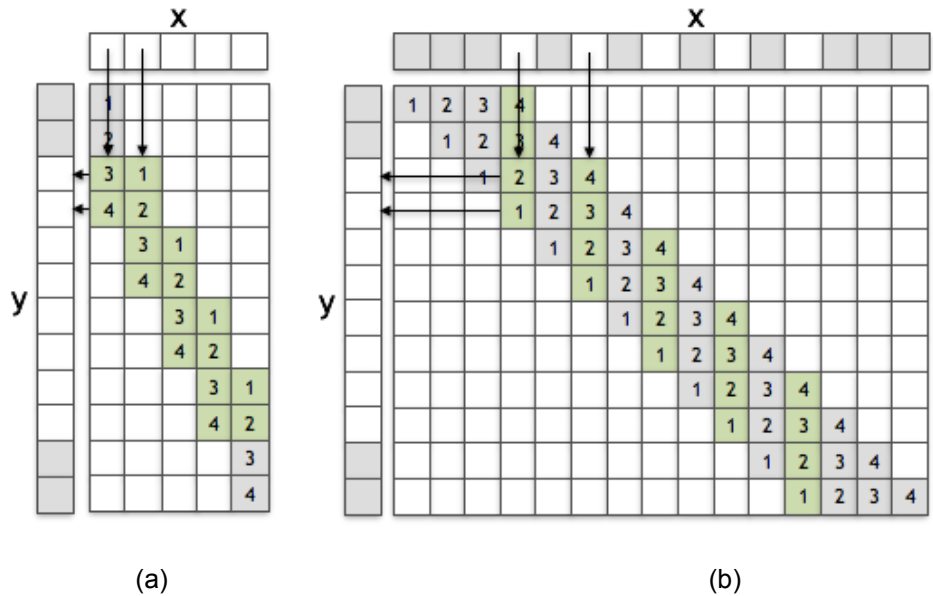

Figure 3: (a) Transposed convolution with stride 2 and (b) sub-pixel convolution with stride ½ in 1D

Fig.3 illustrates 1D cropped transposed convolution and 1D sub-pixel convolution. Both are upsampling operations. In this case, $x$ is of size 5, $f$ is of size 4 and $y$ is of size 8. The grey areas in $y$ represent cropping. The transposed convolution got its name because the matrix for the operation in the middle is a transposed version of the matrix in Fig.2. It is also called backward convolution since it is the backward propagation of a convolutional layer. It is noticeable that the padded convolution becomes a cropped convolution because of the transposed matrix, whereas the sub-pixel convolution got its name from the imaginary sub-pixels with fractional indices filled in between the original pixels. We can see that the only difference between these two operations is that the indices of the weights used when contributing to $y$ from $x$ are different. If we reverse the element indices of filter $f$ in the sub-pixel convolution then this layer will be identical to a transposed convolution layer. In other words, both operations can achieve the same result if the filter is learned.

**Section 2: Deconvolution layer vs Convolution in LR**

In this note, we also want to demonstrate that a simple convolutional layer with kernel size $(o * r^2, i, k, k)$ - e.g. (output channels, input channels, kernel width, kernel height) in LR space is identical to a deconvolution layer with kernel size $(o, i, k * r, k * r)$ where k is a positive integer. We will do this in 2D so the reader can relate it to Fig.1. To avoid overly complicated figures, let's start with a simple sub-pixel padded convolutional layer with a $(1, 4, 4)$ input and a $(1, 1, 4, 4)$ kernel, and assume an upscaling factor of 2 leading to a $(1, 8, 8)$ output:

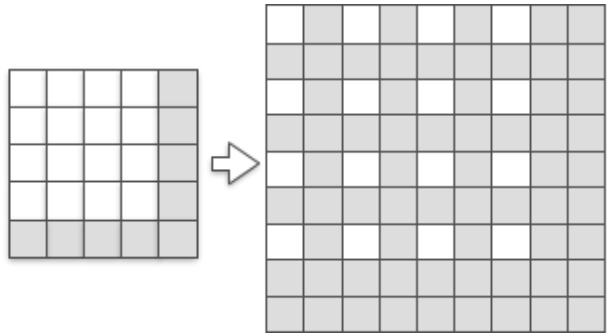

Figure 4: Step 1 of sub-pixel convolution: create sub-pixel image from LR image

As in the 1D case, we create a sub-pixel image with fractional indices from the original input, where the white pixels are the original LR pixels and the grey ones are the zero padded imaginary sub-pixels.

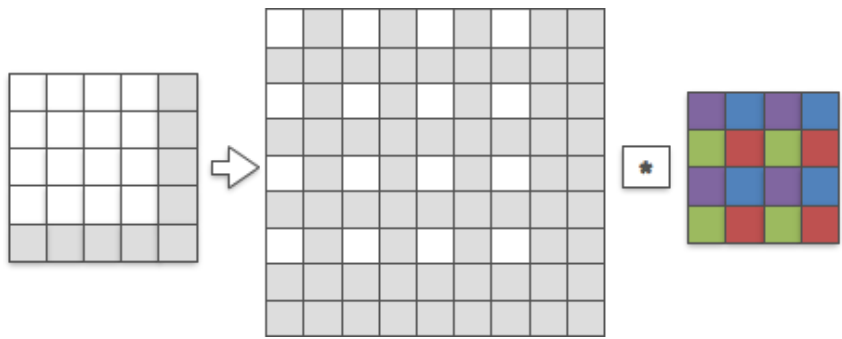

Figure 5: Step 2 of sub-pixel convolution: convolution in sub-pixel space

If a $(1, 1, 4, 4)$ kernel is convolved with the sub-pixels, the first set of weights that are activated by non-zero pixels are the purple ones. Then we move one sub-pixel to the right in the sub-pixel image and the blue weights are activated. Same goes for the green and the red ones.

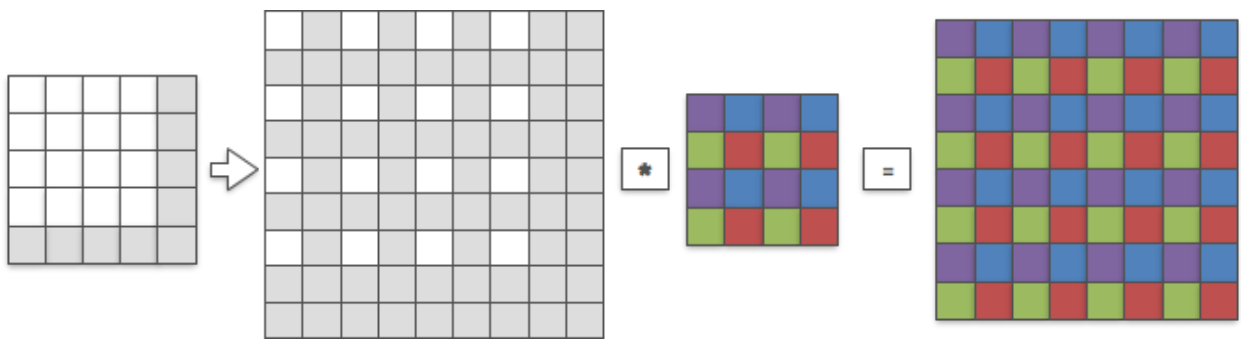

Figure 6: Full view of the sub-pixel convolution

Finally, the output HR image has the same dimension as the sub-pixel image, we color code it to show which set of weights contributed to the pixel.

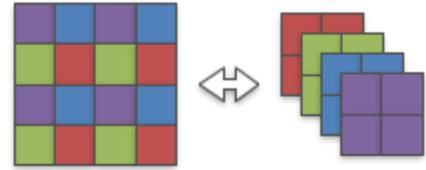

We notice that the different sets of weights in the (1, 1, 4, 4) kernel are activated independently from each other. So we can easily break them into (4, 1, 2, 2) kernels as shown in the figure on the right. This operation is invertible because each set of the weights are independent from each other during the convolution.

In our paper, instead of convolving the (1, 1, 4, 4) kernel with the unpooled sub-pixel image, we convolve the (4, 1, 2, 2) kernel with the LR input directly as illustrated by the following figure:

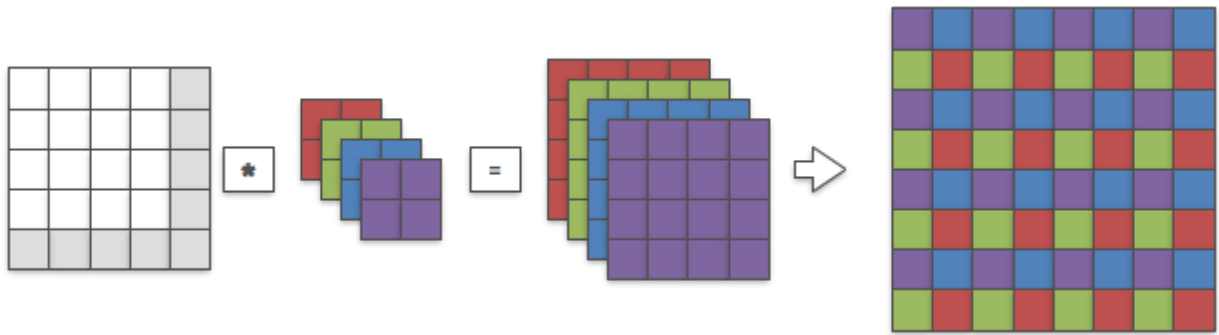

Figure 7: Full view of the proposed sub-pixel convolution using just convolution

When we get the (4, 4, 4) output, we can simply use the periodic shuffling operation mentioned in our paper to reshape the output channels to the HR output. The result is then identical to the HR output in Fig.6. It generalizes to any kernel shape of dimension $(o, i, k, k)$ and rescale ratio $r$. We will leave this as an exercise to the reader.

Here's the trained last convolutional layer kernels of size (9, 32, 3, 3) from our paper on top and the recreated deconvolution layer kernels of (1, 32, 9, 9) on bottom using the inverse operation illustrated on the right:

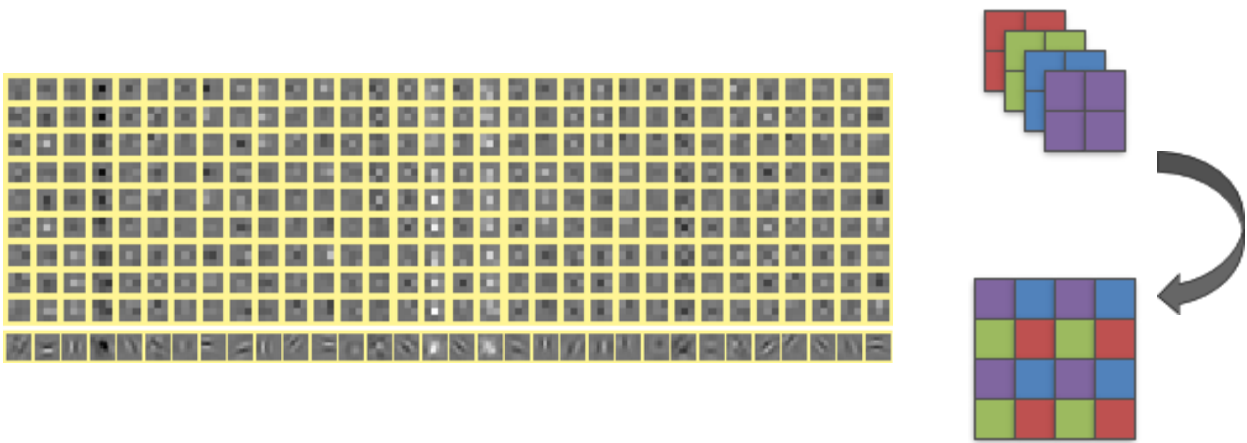

If we apply the top convolutional layer kernels to any 32 channel inputs followed by periodic shuffling we will get the same result as if we apply the bottom deconvolution layer with the $(1, 32, 9, 9)$ kernels. Going back to our 1D case in Fig.3, we simply replace $f = (1,2,3,4)$ with $f_1 = (2,4)$ and $f_2 = (1,3)$, produce $y_1 = f_1 * x$ and $y_2 = f_2 * x$ where $*$ denotes convolution, then combine $y_1$ and $y_2$ to create $y$. The equivalence between convolution in LR space and sub-pixel convolution discussed above applies to $f$ with size equals to $k*r$. But reader might has noticed that for sub-pixel convolution, $f$ can be of any size. However, convolution in LR space actually also works for $f$ with size not equals to $k*r$. For example, if $f = (1,2,3)$ then we will simply have $f_1 = (2)$ and $f_2 = (1,3)$, produce $y_1 = f_1 * x$ and $y_2 = f_2 * x$, then combine $y_1$ and $y_2$ to create $y$.

**Section 3: What does this mean?**

In conclusion, the deconvolution layer is the same as the convolution in LR with $r^d$ channel output where $d$ is the spatial dimension of the data. This means that a network can learn to use $r^2$ channels of LR image/feature maps to represent one HR image/feature maps if it is encouraged to do so. And the operation used to create the $r^2$ channels is just simple convolutions which is no different from the operation used to create the $n_{l-1}$ feature maps before it.

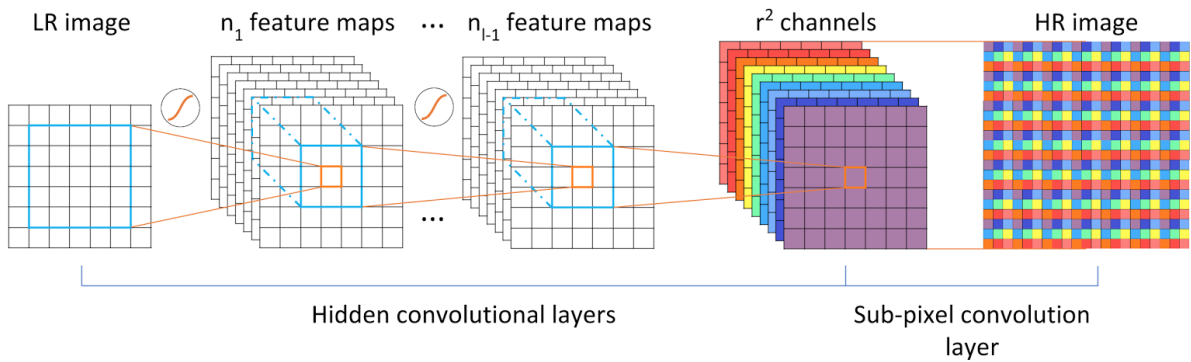

Here comes the additional insights to the problem we have gained during last year after we finished the paper, if we now focus on the convolutional layers before the last convolution, which has $n$ feature maps, we now know that with an upsampling factor of 2 it can learn to represent $n$ feature maps in LR space that are equivalent to $\frac{n}{4}$ feature maps in HR space. Now imagine two networks with the same run-time speed. One has $n = 32$ feature maps all in LR space (LR network) as in [1] and another network upsamples before convolution as in [10] and has $\frac{n}{4} = 8$ feature maps all in HR space (HR network). The representation power of the LR network is actually greater than the HR network at the same run-time speed.

To be more specific, for the LR network, the complexity of the network $O(32 \times 32 \times 3 \times 3 \times \frac{W}{2} \times \frac{H}{2})$ is the same as the HR network $O(8 \times 8 \times 6 \times 6 \times W \times H)$ where $W$ and $H$ denote width and height of the images. The information retained in the feature maps are also the same between the LR $(l \times 32 \times \frac{W}{2} \times \frac{H}{2})$ and the HR $(l \times 8 \times W \times H)$ network, where $l$ denotes the number of layers. The receptive fields of each activation is equivalent in the original input LR space. However, the number of parameter of the LR network $(l \times 32 \times 32 \times 3 \times 3)$ is larger than that for the HR network $(l \times 8 \times 8 \times 6 \times 6)$. Thus the network with convolutions exclusively in LR has more representation power than a network that upsamples the input at the same speed.

Given the above argument, we now think that for super-resolution problems, an explicit upsampling using a bicubic or a deconvolution layer isn't really necessary. For example, independently developed later works by Dong [12] and Johnson [13] use convolution in LR for super resolution and even style transfer.

This raises more interesting questions. Is explicit upsampling using bicubic interpolation or deconvolution necessary in any other applications? Can the network learn when to upscale and what percentage of feature maps to upscale from using only convolutions? What happens when resNet is combined with many layers of convolutions for tasks which require upsampling, will the network learn to combine LR and HR features automatically? We will leave the readers to ponder these more interesting questions.